\documentclass{iopjournal}

\usepackage[numbers,sort&compress]{natbib}
\usepackage{doi}
\usepackage{ragged2e}
\usepackage{lmodern}

\usepackage{amsmath}

\usepackage{arydshln}
\usepackage{colortbl}
\definecolor{rulegrey}{gray}{0.5}
\definecolor{dashgrey}{gray}{0.6}
\usepackage{soul,xcolor}

\usepackage{graphicx}
\usepackage{silence}
\usepackage{subcaption}
\DeclareCaptionFont{iopcap}{\fontsize{8}{10}\selectfont}
\usepackage{subcaption}

\usepackage[inline]{enumitem}
\newlist{inenum}{enumerate*}{1}
\setlist[inenum]{%
	label=(\arabic*),          
	labelsep=0.4em,            
	itemjoin={{; }},           
	itemjoin*={{; and }},      
	leftmargin=0pt,            
	itemsep=0pt,               
	topsep=0pt, 
	parsep=0pt
}

\usepackage{booktabs}
\usepackage{multirow}
\usepackage[table]{xcolor}
\usepackage{threeparttable}
\usepackage{makecell}

\usepackage[nameinlink]{cleveref}
\crefname{figure}{Figure}{Figs.}
\crefname{equation}{Equation}{Eqs.}

\newcommand*{\TallestContentZero}{$x_1$}
\newcommand*{\rax}[1]{\overrightarrow{\makebox[1.25em]{$#1$\vphantom{\TallestContentZero}}}}

\begin{document}
\articletype{Paper}
\setcitestyle{square}

\title{Defining Energy Indicators for Impact Identification on Aerospace Composites: A Structured Feature Selection Approach Guided by Domain Knowledge}

\author{Nat\'{a}lia Ribeiro Marinho$^{1,*}$\orcid{0000-0002-1042-3571}, Richard Loendersloot$^1$\orcid{0000-0002-1113-8203}, Jan Willem Wiegman$^2$\orcid{0000-0002-1066-7234}, Frank Grooteman$^2$\orcid{0000-0001-8036-3845}, Uraz Odyurt$^1$\orcid{0000-0003-1094-0234} and Tiedo Tinga$^1$\orcid{0000-0001-6600-5099}}

\affil{$^1$Engineering Technology Faculty, Department of Mechanics of Solids, Surfaces  and  Systems, Engineering Dynamics (ED), University of Twente (UT), Enschede, NL}

\affil{$^2$Department of Aerospace Vehicles Integrity and Life Cycle Support (AVIL), Royal Netherlands Aerospace Centre (NLR), Marknesse, NL}

\affil{$^*$Author to whom any correspondence should be addressed.}

\email{n.ribeiromarinho@utwente.nl}


\keywords{impact energy estimation; structural health monitoring; aerospace composites; feature extraction and selection; input space design; domain knowledge; data representation}

\begin{abstract}
\justifying
Energy estimation is critical to impact identification on aerospace composites, where low-velocity impacts can induce internal damage that is undetectable at the surface. Data sparsity, signal noise, complex feature interdependencies, non-linear dynamics, massive design spaces, and the ill-posed nature of the inverse problem often constrain current methodologies for energy prediction. Machine learning enriched with prior knowledge is a promising direction for overcoming these constraints. Prior knowledge can be incorporated by acting on the input space, where the choice of data representation directly influences how effectively the model relates measured signals to impact energy. Despite its importance, the selection of effective features lacks a systematic procedure, with no consensus on how to choose among the many candidate descriptors available. The present study addresses that gap through a structured workflow that designs the input space using domain knowledge. Features are extracted from the time, frequency, and time-frequency domains, then filtered for statistical significance, correlation, dimensionality reduction, and robustness to noise. Exploratory data analysis further relates the retained descriptors to the dynamics, yielding a reduced group of features that captures essential mechanisms such as amplitude scaling, spectral redistribution, and transient signal behaviour. The resulting indicators form the input space for a fully connected neural network, which is trained and validated on experimental data from multiple impact scenarios, including pristine and damaged states. The model reduces the prediction error by a factor of three relative to conventional time-series techniques and purely data-driven baselines, while every retained or discarded descriptor remains traceable to the aspect it describes. Overall, the framework advances predictive performance, interpretability, and diagnostic confidence by embedding domain knowledge through targeted feature selection.
\end{abstract}


\justifying

\section*{Introduction}
\label{sec:introduction}

Aerospace composite structures are susceptible to Barely Visible Impact Damage (BVID), which poses a critical threat to both structural integrity and operational safety. These internal damages are typically undetectable by conventional visual inspection and can significantly degrade load-bearing capacity~\cite{Seno2020A, DeSimone2017Impact}. To mitigate these risks, Structural Health Monitoring (SHM) systems have been developed to capture dynamic responses using embedded sensors. These systems enable estimation of impact energy and detection of internal damage, supporting timely maintenance, enhancing operational reliability, and informing design improvements~\cite{Rezazadeh2025Systematic, ZHU2020103508, YANG2021107186, Andreades2020Tensile}.

A central challenge in this context is reconstructing impact energy from measured responses, which is an ill-posed inverse problem. The relationship between sensor signals and unknown impact conditions is typically non-unique, highly sensitive to noise, and underdetermined. These difficulties intensify under practical constraints, such as sparse sensor layouts, uncertainties in measurements, and non-linear structural behaviour at high excitation levels~\cite{Seno_2021}. Consequently, conventional methods frequently produce poor estimates unless constrained by regularisation or supported by computationally intensive formulations~\citep{ZHOU2024111238, LIU2023107873, DeSimone2019A}. To ensure reliable performance under these conditions, estimation strategies must effectively manage incomplete and noisy data, preserve numerical stability, and support real-time execution within embedded systems.

Research efforts have addressed these challenges using numerical, analytical, data-driven, and probabilistic approaches. In the numerical and analytical domains, \citet{LIU2023107873} developed a finite-element model for woven laminates subjected to low-velocity impact. Their model reconstructed force histories with reasonable accuracy but required intensive computation and relied exclusively on simulated virtual accelerations. \citet{Correas2021Analytical} introduced an analytical spring-mass model for stiffened panels that is in strong agreement with numerical simulations. However, the model employed simplified loading assumptions, idealised boundary conditions and lacked experimental validation. \citet{Seydel2001Impact} reconstructed impact location and force history in real time on stiffened panels using a smoother-filter algorithm, with an accuracy that is limited to the boundary conditions assumed in the model.

Data-driven methods have also been employed to model complex and non-linear relationships between structural responses and unknown impact conditions. \citet{tabian2019convolutional} employed a Convolutional Neural Network (CNN) to localise and quantify impacts on composite structures. Similarly, \citet{sharif2013smart} successfully reconstructed impact force histories from time-series data, and \citet{Zhu2023Impact} assessed impact energy in sandwich composites through an ensemble deep learning approach. \citet{Zhao2025Impact} combined a convolutional network for localisation with a temporal network for force reconstruction, reducing data demand through domain-invariant features, with validation so far limited to a simplified structure and a low-energy range. Although these approaches are effective in controlled environments, they typically require large, high-quality datasets and struggle to generalise to new or noisy scenarios~\citep{Damm2020Deep}. Their dependence on data representativeness and low interpretability further constrained practical deployment.

Additional developments have explored probabilistic techniques. \citet{yan2017impact} introduced a Bayesian regularisation framework, combined with an unscented Kalman filter, to improve robustness to noise in impact identification. Although the method improved stability, it remained sensitive to sensor configuration and limited to low-energy regimes. \citet{Huang_2023} employed transfer learning in a deep learning model, reaching acceptable accuracy but requiring extensive tuning and access to reference force histories. \citet{ZHANG2020111882} integrated experimental testing with high-fidelity simulations on honeycomb panels, yielding detailed insights but incurring high computational costs and sensitivity to parameterisation. Despite their contributions, probabilistic methods remain computationally demanding, difficult to scale, and dependent on sensor configurations and reference data, which limits their practical adoption.

Across the aforementioned modelling strategies, no single paradigm resolves every challenge encountered in practical SHM applications. Numerical and analytical models include explicit prior knowledge of the structure, yet apply only under the assumptions used to build them~\cite{Colombo2021Shape}. Data-driven models capture complex relationships without such assumptions, but depend on representative and high-quality training data and remain sensitive to the variability of impact conditions~\cite{Momeni2022Highdimensional, Seno2019Passive}. Probabilistic formulations represent an intermediate solution that combines prior knowledge with sensor data; however, their accuracy demands substantial effort to tune and regularise the model, and depends further on the degree of informative prior integration achieved~\cite{Giannakeas2021Digital, Ni_2020, Hughes_2021}. Across all three, the ill-posed nature of the inverse problem persists, since regularisation becomes costly on large problems~\cite{Qiao2017Sparse} and the identification may stay underdetermined when measurements are sparse~\cite{Liu2021Impact}.

In response to these limitations, a promising direction combines the strengths of physics-inspired assumptions and data-driven methods, retaining domain knowledge in analytical and numerical models while benefiting from machine learning algorithms' ability to capture non-linear relationships between measured response and impact energy. To date, most work in this direction has concentrated on the model itself, through architectures and training strategies that embed such knowledge. The input to that model has received comparatively less attention, even though it determines which information reaches the algorithm in the first place.

In cases where the input has been addressed, the reported choices for data representation in impact identification vary widely. Some methods work on the raw time signal, using arrival times or transfer functions to locate the impact and reconstruct the force~\citep{DeSimone2017Impact, DeSimone2019A}. Others adopt spectral components to capture differences in energy distribution and attenuation that are less apparent in the raw signal~\citep{ghajari2013identification, Wu2015Impact}. A third group applies time-frequency representations, including wavelet components, to capture the non-stationary and dispersive character of propagating waves due to interactions with complex components~\citep{Qiu2011A}. More recent work applies deep learning to derive representations directly from the signal and combines them with machine learning mechanisms~\citep{Huang_2023, Zhao2025Impact}.

This heterogeneity in approaches to defining the input space has practical consequences. Each representation emphasises a different part of the impact response and addresses a specific difficulty, such as anisotropy, sparse sensing, or operational variability. As a result, methods differ in their data requirements, sensitivity to operating conditions, and necessary sensor layouts. Direct comparison, therefore, becomes difficult, and reported performance remains confined to the structure and objective under study. These previous investigations also cover only part of the impact identification pipeline, concentrating mainly on source localisation and damage characterisation. By contrast, relatively few existing publications target energy estimation, and so there is no established systematic procedure for feature selection in impact energy estimation using sensor data. This study addresses that gap by treating the design of the input space as an explicit step in the monitoring framework. Indeed, a structured sensor data representation, one whose selected inputs reflect the relevant impact dynamics, is expected to improve the reliability of the estimate under noisy and sparse conditions~\citep{FINK2026112376}. A method to define an optimised input space therefore becomes more than a preprocessing detail.


The input space design is guided by domain knowledge, offering a simple and efficient approach that captures the essential dynamical behaviour of the system without added complexity~\citep{khalid2024advancements, kalhori2025advances}. The design method builds on the principle that measurement signals carry information about the input excitation~\citep{muir2021damage, marinho2025evaluating}, spread across the time, frequency, and time-frequency domains. Since each domain contributes distinct and complementary information about the impact, prior knowledge of structural dynamics motivates representing and balancing all three within the input space. Together, they yield a complete characterisation of the impact event that no single domain can achieve.

Although prior studies in acoustic emission and impact analysis~\citep{moevus2008analysis, sibil2012optimization, sharif2018impact, pineau2012subsampling} confirm the diagnostic value of multi-domain descriptors, they also demonstrate their sensitivity to propagation path effects, attenuation, non-linearities, and noise. Given these drawbacks, the framework applies targeted feature selection to retain only the most relevant and robust descriptors. Each candidate is assessed for its sensitivity to impact energy, its redundancy against the others, and its stability under measurement noise. Combining these three aspects into a composite score yields a compact, balanced set of energy-sensitive features that is both informative and efficient. Because every retained feature carries this score, the selection remains traceable throughout the process. The refined feature subset forms the optimised input space for a Fully Connected Neural Network (FCNN) trained to estimate impact energy. Taken together, the central contribution therefore lies not in the individual features but in a selection that represents the input. Hence, the training data reflect the dynamical behaviour of the impact and meet the requirements of SHM under imperfect measurement conditions.

\paragraph*{Use-case}
The procedure is applied to a dataset obtained from intermediate-mass impact tests on a square composite panel, covering multiple impact locations and energy levels. The dataset replicates conditions typical of in-service SHM applications, characterised by low signal quality, limited data availability, and loading scenarios both below and above the damage onset. This configuration provides a realistic benchmark for assessing feature robustness and verifying the effectiveness of the proposed input space design.

\paragraph*{Contribution}
%
The main contributions of this study are:
\begin{itemize}
\item A structured workflow for feature selection;
\item Interpretability and transparency of the impact energy estimation based on ranked scores;
\item Improved robustness under noisy measurement conditions;
\item A compact and computationally efficient input space design;
\item A transferable methodology for input space design for structural dynamics applications.
\end{itemize}

This paper is structured as follows. 
\Cref{sec:selection_method} presents the feature extraction and selection method, detailing each stage. \Cref{system_description} presents the experimental and numerical datasets used for feature evaluation and model validation. \Cref{sec:results} reports the results and assesses model performance. Finally, \Cref{sec:conclusion} summarises the key findings, acknowledges limitations and identifies directions for future research.

\section{Feature extraction and selection method}
\label{sec:selection_method}

\begin{figure}[htbp]
    \centering
    \includegraphics{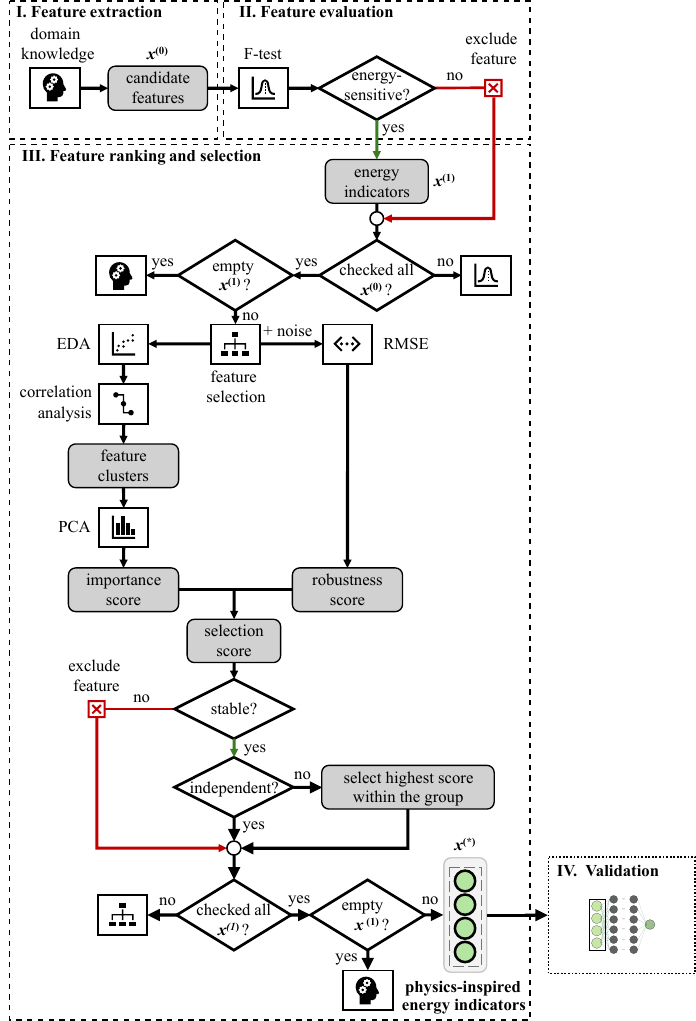}
    \caption{Method for defining an input space guided by domain knowledge. Feature candidates \textit{x\textsuperscript{(0)}}, energy-sensitive features \textit{x\textsuperscript{(1)}}, and physics-inspired energy indicators \textit{x\textsuperscript{(*)}}. Black arrows: processing paths; red arrows: discarded descriptors; and green arrows: selected features.}
    \label{fig:conceptual_framework}
\end{figure}

The proposed method proceeds through a sequence of steps, each addressing a specific property of the descriptors: sensitivity to impact energy, independence, importance and robustness to noise. \Cref{fig:conceptual_framework} presents the complete overview, and the remainder of this section discusses each step in detail. Throughout, the term \textit{input space} refers to the selected descriptors extracted from the sensor signals and used as input to an FCNN for impact energy prediction.

In the first step, candidate features \textit{x\textsuperscript{(0)}} are extracted to capture the dynamic behaviour of the structure under impact. This initial feature set is grounded in structural dynamics and composite material behaviour. Each descriptor is therefore expected to vary with impact energy as the structural response does, a behaviour confirmed later through exploratory data analysis. Features are derived from the time, frequency, and time–frequency domains to represent complementary aspects of the response.

The second step evaluates whether candidate features \textit{x\textsuperscript{(0)}} respond to variations in impact energy by testing their statistical sensitivity using an F-test. Features that do not show significant variation are discarded, leaving a reduced subset of energy indicators \textit{x\textsuperscript{(1)}}. A feedback loop addresses cases where no features pass the statistical test by returning to the extraction stage, where domain knowledge is applied to define new descriptors. The method assumes that sensors capture relevant impact responses. If repeated iterations still fail to identify energy-sensitive features, this may suggest limitations in the sensing system, in which case adjustments to sensor type, configuration, or placement could be considered.

The third step begins with Exploratory Data Analysis (EDA), which examines how each energy indicator behaves across the range of impact energies. Two properties guide the assessment: monotonicity, the degree to which a descriptor changes in a single direction, and trendability, the consistency of that change across energy levels. These trends test the expectations formulated during extraction, since they can be compared against the dynamics anticipated as excitation increases, such as amplitude scaling, saturation under non-linear response, and spectral redistribution. The assessment remains qualitative, yet it is decisive for judging whether the descriptors behave as the dynamics suggest. 

Based on these insights, correlation analysis and Principal Component Analysis (PCA) are applied in parallel to reduce dimensionality and eliminate redundancy. Correlation analysis groups descriptors with strong dependencies, while PCA quantifies the contribution of each feature to overall variability, providing importance scores that guide ranking and selection.  This step assumes signals of sufficient quality, since PCA maximises variance regardless of its origin. In the presence of severe noise, components may reflect noise rather than the structural response. If this occurs, adjustments to the acquisition setup or improved signal processing strategies could be considered.

To assess robustness to measurement noise, white Gaussian noise is added to the extracted features to simulate low-quality signals, following previous impact identification studies~\cite{Sharif-Khodaei_2012}. Although limited to white noise, the perturbation is controlled and therefore sufficient to compare feature stability during selection. The influence of noise is quantified by the Root Mean Square Error (RMSE) between the original and perturbed values, which is then used as a robustness score. This score is then combined with the PCA importance score to yield a selection score that balances relevance with consistency under uncertainties in measurements. Within each correlation cluster, only the feature with the highest selection score is retained, resulting in a final set of physics-inspired energy indicators \textit{x\textsuperscript{(*)}} that are relevant, independent, and reliable under noisy conditions. If no features meet these criteria, the process returns to the extraction stage for refinement based on updated domain knowledge.

In the final step, the identified physics-inspired energy indicators are utilised to train an Artificial Neural Network (ANN) for estimating impact energy. Once trained, the ANN employs inputs from the test dataset to generate predictions. Predictive performance is assessed under sparse and unbalanced conditions by comparing estimated values against experimental ground truth. This validation reflects conditions typically encountered in real-world applications.

The following sections describe each stage of the method in detail, with an emphasis on how domain knowledge shapes the input space for impact identification. The method does not rely on impact-specific processing steps, which makes it applicable beyond this case study. Consequently, it provides a transferable methodology for feature selection in other structural dynamics applications.

\subsection{Feature extraction}
\label{subsec:energy_indicators}

The candidate features denoted as \( x^{(0)} \) in Step I of \Cref{fig:conceptual_framework} form a multi-domain set of descriptors extracted from sensor signals during drop-weight impact tests. The complete list of candidate features is shown in \Cref{tab:feature_extraction}, which compiles features reported in prior studies as effective for characterising structural dynamics from time-series measurements. Their inclusion in the present framework is motivated by criteria specific to impact loading: features that can correlate with impact severity, capture the transient and localised nature of impact events, reflect the material behaviour, and remain measurable with practical sensor layouts.

The final columns identify which mechanisms each descriptor may capture. Four mechanisms are considered, since the recorded signal relates to the impact energy through several distinct routes. At low energy, the response remains elastic, so the amplitude and energy of the signal linearly increase with the impact energy; amplitude scaling (AS) is therefore included as the first mechanism~\cite{Garcia2019, Wu2015Impact, He2024}. At higher energy, this proportionality no longer holds, since part of the energy is absorbed by non-linear behaviour; saturation under non-linear response (SN) is therefore included as the mechanism marking the limit of the linear elastic relation~\cite{Feraboli2006}. Increasing impact severity may also shift energy across the spectrum, so spectral redistribution (SR) is included as a mechanism independent of amplitude~\citep{Gao2020, Barile2021}. Changes in the damage state (DS) of the structure are expected to lower the modal frequencies, may raise the high-frequency content, and reduce the amplitude, so sensitivity to that state is included as the fourth mechanism~\citep{Alnuaimi2020, Ciardiello2025}. A descriptor is admitted as a candidate when it can capture at least one mechanism; whether each descriptor follows the expected behaviour is examined through exploratory data analysis in \Cref{sec:ranking_selection}.
\begin{table}[htbp]
  \setlength{\intextsep}{0pt}
  \centering
  \small
  \setlength{\tabcolsep}{4pt}
  \hyphenpenalty=10000
  \exhyphenpenalty=10000
  \newcolumntype{M}[1]{>{\raggedright\arraybackslash}m{#1}}
  \newcolumntype{C}[1]{>{\centering\arraybackslash}m{#1}}
  \caption{Candidate features, their description, and the mechanisms through which each descriptor may relate to the impact event.}
  \label{tab:feature_extraction}
  \begin{tabular}{@{}M{1.6cm}M{1cm}M{2.5cm}M{3cm}C{0.9cm}C{0.9cm}C{0.9cm}C{0.9cm}M{1cm}@{}}
    \toprule
    \multirow{2}{*}{\textbf{Domain}} & \multirow{2}{*}{\textbf{ID}} & \multirow{2}{*}{\textbf{Feature}} & \multirow{2}{*}{\textbf{Description}} & \multicolumn{4}{c}{\textbf{Link with impact dynamics}} & \multirow{2}{*}{\textbf{Ref.}} \\
     & & & & \textbf{AS} & \textbf{SN} & \textbf{SR} & \textbf{DS} & \\
    \midrule
    \arrayrulecolor{dashgrey}
    \textbf{Time} & PA & Peak amplitude & Maximum signal amplitude & $\bullet$ & $\bullet$ & & & \cite{ghajari2013identification,Feraboli2006} \\
    \cdashline{2-9}
     & TE & Transmitted energy & Area under signal envelope & $\bullet$ & $\bullet$ & & & \cite{tabian2019convolutional,Wu2015Impact} \\
    \cdashline{2-9}
     & RT & Rise time & Time from signal onset to peak amplitude & $\bullet$  & & & & \cite{doan2015unsupervised,Feraboli2006} \\
    \cdashline{2-9}
     & CTP & Counts to peak & Number of counts from onset to peak amplitude & $\bullet$ & & & & \cite{kostopoulos2003identification,farrar2012structural} \\
    \cdashline{2-9}
     & RA & Rise angle & PA / RT & $\bullet$ & & & & \cite{wisner2019acoustic,Cortes2018} \\
    \cdashline{2-9}
     & RMS & RMS & Root mean square & $\bullet$ & & & & \cite{jang2015impact,Cortes2018} \\
    \cdashline{2-9}
     & EPR & Energy peak ratio & TE / PA & $\bullet$ & $\bullet$ & & & \cite{guel2020data,Cortes2018} \\
    \cdashline{2-9}
     & NDA & Non-dimensional amplitude & PA / mean signal amplitude & $\bullet$  & & & & \cite{wisner2019acoustic,roberts2013computational} \\
     \cmidrule[0.2pt](l{-4pt}r{-4pt}){1-9}
    \arrayrulecolor{dashgrey}
    \textbf{Frequency} & CF & Centroid frequency & PSD centre of gravity & & & $\bullet$ & & \cite{sultan2011impact,farrar2012structural} \\
    \cdashline{2-9}
     & PF & Peak frequency & Frequency with maximum power & & & $\bullet$ & & \cite{kostopoulos2003identification,VanSteen2021} \\
    \cdashline{2-9}
     & WPF & Weighted peak frequency & $\sqrt{\text{PF} \cdot \text{CF}}$ & & & $\bullet$ & & \cite{ali2019microscopic,VanSteen2021} \\
    \cdashline{2-9}
     & PCR & Peak centroid ratio & PF / CF & & & $\bullet$ & & \cite{muir2021damage,VanSteen2021} \\
    \cdashline{2-9}
     & RON & Roll-on frequency & Frequency at which 10\% of total PSD has accumulated & & & $\bullet$ &  & \cite{guel2020data,farrar2012structural} \\
    \cdashline{2-9}
     & ROFF & Roll-off frequency & Frequency at which 90\% of total PSD has accumulated & & & $\bullet$ & & \cite{guel2020data,farrar2012structural} \\
    \cmidrule[0.2pt](l{-4pt}r{-4pt}){1-9}
    \arrayrulecolor{dashgrey}
    \textbf{\makecell[tl]{Time-\\frequency}}  & AM & Approximation max & Maximum value of low-frequency WPT component (level 3) & $\bullet$ & $\bullet$ & & & \cite{oskouei2009wavelet,Li2021AE} \\
    \cdashline{2-9}
     & DM & Detailed max & Maximum value of high-frequency WPT component (level 3) & $\bullet$ & & & $\bullet$ & \cite{oskouei2009wavelet,Li2021AE} \\
    \cdashline{2-9}
     & AME & Approximation max energy & Energy of AM component & $\bullet$ & $\bullet$ & & & \cite{fotouhi2015investigation,Li2021AE} \\
    \cdashline{2-9}
     & DME & Detailed max energy & Energy of DM component & $\bullet$ & & & $\bullet$ & \cite{fotouhi2015investigation,roberts2013computational} \\
    \arrayrulecolor{black}
    \bottomrule
    \addlinespace
    \multicolumn{9}{@{}l@{}}{\textbf{AS}: amplitude scaling in the elastic regime; \textbf{SN}: saturation under non-linear response} \\
    \multicolumn{9}{@{}l@{}}{\textbf{SR}: spectral redistribution; \textbf{DS}: changes in the damage state} \\
    \multicolumn{9}{@{}l@{}}{\textbf{PSD}: Power Spectrum Density; \textbf{WPT}: Wavelet Packet Transform} \\
  \end{tabular}
\end{table}

The reason for grouping features into time, frequency, and time–frequency domains is that each captures a distinct but complementary aspect of the structural response. Time-domain descriptors quantify impact intensity and energy dissipation, linking directly to the overall energy content~\citep{tabian2019convolutional, cuomo2023damage, davies2004impact}. Frequency-domain features characterise the spectral distribution and identify dominant frequency components of impact events, reflecting the dynamic behaviour of the target structure~\citep{sharif2018impact, ohayon1997structural}. Time-frequency features capture energy distribution across multiple frequency bands, integrating both transient high-frequency components and low-frequency structural responses~\citep{oskouei2009wavelet, fotouhi2015investigation}. No single domain covers all four mechanisms, so integrating these three perspectives yields a complete characterisation of the event, overcomes the limitations of single-domain approaches and increases sensitivity to both global and localised effects. Representing all three domains is also expected to improve performance where data are limited and imperfect, since embedding domain knowledge at the feature level supplies prior information and supports generalisation under sparse and noisy conditions~\citep{hao2022physics, karniadakis2021physics}. 

\subsection{Feature evaluation}
\label{subsec:ANOVA_methods}

The feature evaluation stage investigates whether candidate features exhibit statistically significant sensitivity to variations in impact energy. Impactor mass, impactor diameter, and impact location directly influence the amount and distribution of transferred energy~\citep{LI2021122491,YANG2020112027,LIAO2020105783}, so these factors were included together with energy as the main parameters controlling the impact response. An F-test within an Analysis of Variance (ANOVA) framework was applied to quantify feature sensitivity while accounting for variations introduced by these parameters. The parameters and their levels were:
\begin{enumerate}[label=\roman*]
	\item \textbf{Impact Energy:} Two levels (2~J, 20~J) to represent distinct loading conditions;
	\item \textbf{Impactor Diameter:} Two levels (16~mm, 50~mm) to assess variations in stress distribution and contact area;
	\item \textbf{Impactor Mass:} Two levels (0.5~kg, 2~kg) to account for differences in momentum transfer;
	\item \textbf{Impact Location:} Two positions (IC2 and IC4, \Cref{fig:subfig_setup}) to assess structural variations and attenuation effects.
\end{enumerate}

The selected factors and their two-factor interactions were incorporated into a linear regression model to predict the response variable (i.e., candidate features), following the approach of \citet{montgomery2017design}:
\begin{equation}
	y_n =\beta_0 + \sum_{i=1}^{p} \beta_i x_i + \sum_{i=1}^{p-1} \sum_{j=i+1}^{p} \beta_{ij} x_i x_j + \epsilon_n \texttt{,}
	\label{eq:regression_model}
\end{equation}
where $x_i$ and $x_j$ are the coded numeric values representing the four primary factors: impact energy, impactor diameter, impactor mass, and impact location. The parameter $\beta_0$ indicates the average response across all test runs. The coefficients $\beta_i$ quantify individual factor effects, while $\beta_{ij}$ measures the combined influence of two factors. Additionally, the term $\epsilon_n$ denotes a random error modelled as a normal distribution with a zero mean. 

The statistical model described in \Cref{eq:regression_model}  allows for an analysis of variance to test the null hypothesis that either $\beta_i = 0$ or $\beta_{ij} = 0$. The test statistic is based on the expected values of the mean square error ($\varepsilon_\text{MSE}$) and the mean square regression ($\sigma_\text{MSR}$). In ANOVA, this measure is represented by the F-score as
\begin{equation}
	F_{calc} = \frac{\varepsilon_\text{MSE}}{\sigma_\text{MSR}}\texttt{,}
\end{equation}
where $\varepsilon_\text{MSE}$ and $\sigma_\text{MSR}$ are defined as
\begin{equation}
	\varepsilon_\text{MSE} = \frac{\sigma_b^2}{b} \quad\text{and}\quad \sigma_\text{MSR} = \frac{\sigma_w^2}{w}.
\end{equation}
In this formulation, the term $\sigma_b^2$ represents the variance in feature values between experimental conditions (e.g., energy, mass, diameter, and location), reflecting the influence of structural and loading parameters. In contrast, $\sigma_w^2$ represents the variance within each setting, based on repeated measurements, and accounts for measurement noise or uncontrolled disturbances. The parameters $b$ and $w$ correspond to the number of independent comparisons between test conditions (between-group) and repetitions (error or within-group), respectively. 

Finally, the significance of the means~\citep{fisher1934statistical}, defined here as the feature sensitivity to varying energy levels, was determined by comparing the computed F-score ($F_{calc}$) with the critical F-value ($F_{crit}$), obtained from F-distribution tables at a 5\% significance level. A factor or interaction is considered statistically significant if the F-score exceeds the critical value.

This study used an orthogonal array~\citep{hedayat2012orthogonal} to structure the factor levels with statistical rigour and to represent impact conditions by systematically varying the four factors. The resulting design, represented by the test matrix in \Cref{tab:testmatrix_conf}, was then used to define the confirmation experiments. These confirmation experiments were implemented as simulated impact responses generated according to this design. Synthetic data were required because the experimental dataset lacked the necessary structured variation, whereas the validated simulation model was well-suited to this purpose. Details of the system configuration and numerical dataset are provided later in \Cref{system_description}.
\begin{table*}
	\centering
	\caption{Test matrix for confirmation experiments.}
	\label{tab:testmatrix_conf}
	\begin{tabular}{@{}ccccc@{}}
	  	\toprule
	    \textbf{ID} & \textbf{Energy [J]} & \textbf{Diameter [mm]} & \textbf{Mass [kg]} & \textbf{Location} \\
	    \midrule
	    1     & 2     & 16    & 0.5   & IC4 \\
	    2     & 20    & 50    & 0.5   & IC4 \\
	    3     & 2     & 50    & 2     & IC4 \\
	    4     & 20    & 16    & 2     & IC4 \\
	    5     & 2     & 50    & 0.5   & IC2 \\
	    6     & 20    & 16    & 0.5   & IC2 \\
	    7     & 2     & 16    & 2     & IC2 \\
	    8     & 20    & 50    & 2     & IC2 \\
	    \bottomrule
	\end{tabular}
\end{table*}

To assess variance significance, the eight confirmation experiments (ID1 to ID8 in \Cref{tab:testmatrix_conf}) were arranged into ANOVA evaluations for the F-test. \Cref{tab:Ftest_analysis} details these evaluations, which focus on the effect of impact energy and examine whether impactor mass, diameter, and location also contribute or interact with energy in shaping the response. For example, ANOVA evaluation \#1 evaluates impact energy at two levels (2~J, 20~J) and impactor diameter at two levels (16~mm and 50~mm), including their interaction, with two repetitions ($n_1$ and $n_2$). In the case of synthetic data, these repetitions arise from repeated factor combinations within the orthogonal array, allowing for balanced sampling and valid statistical comparisons. Overall, this methodology ensures that feature sensitivity to impact energy is evaluated under varying conditions relevant to operational scenarios.
\begin{table*}
	\centering
	\caption{F-test ANOVA evaluations based on combinations from confirmation experiments (ID1 to ID8).}
	\label{tab:Ftest_analysis}
	\begin{tabular}{@{}clllcc@{}}
	    \toprule
	    \textbf{ANOVA evaluation $\#$} & \multicolumn{2}{c}{\textbf{Factor}} & & \textbf{\textit{n}\textsubscript{1}} & \textbf{\textit{n}\textsubscript{2}} \\
	    \midrule
	    \multirow{5}{*}{1} & \textbf{Diameter [mm]} & \textbf{Energy [J]} & & & \\
	    & 16 & 2 & & ID1 & ID7 \\
	    & 16 & 20 & & ID4 & ID6 \\
	    & 50 & 2 & & ID3 & ID5 \\
	    & 50 & 20 & & ID2 & ID8 \\
		\arrayrulecolor{gray!50}
		\midrule
	    \multirow{5}{*}{2} & \textbf{Mass [kg]} & \textbf{Energy [J]} & & & \\
	    & 0.5 & 2 & & ID1 & ID5 \\
	    & 0.5 & 20 & & ID2 & ID6 \\
	    & 2 & 2 & & ID3 & ID7 \\
	    & 2 & 20 & & ID4 & ID8 \\
		\midrule
	    \multirow{5}{*}{3} & \textbf{Location} & \textbf{Energy [J]} & & & \\
	    & IC4 & 2 & & ID1 & ID3 \\
	    & IC4 & 20 & & ID2 & ID4 \\
	    & IC2 & 2 & & ID5 & ID7 \\
	    & IC2 & 20 & & ID6 & ID8 \\
	    \arrayrulecolor{black}
	    \bottomrule
	\end{tabular}
\end{table*}

\subsection{Feature ranking and selection}
\label{subsec:target_engineering}

Following feature evaluation, the feature ranking and selection scheme defines an optimised input space by eliminating redundancies and enhancing interpretability. This process integrates correlation evaluation, Principal Component Analysis (PCA),  and robustness test to retain informative, independent, and stable features.

First, a correlation analysis is conducted to identify and eliminate redundant features. This analysis is essential as redundant features can obscure significant patterns in the data~\citep{blum1997selection}.  Linear dependencies between indicator pairs were assessed using the Pearson correlation coefficient, with values ranging from -1 to +1. A coefficient of -1  means perfect negative correlation, +1 perfect positive correlation, and zero represents no linear relationship between the variables~\citep{sarker2021machine}.  For two random variables, \( S \) and \( Q \), with \( k \) observations, the coefficient is defined as~\citep{han2022data}
\begin{equation}
	p~(S,Q)=\frac{\sum\limits_{i=1}^k{(S_i-\overline{S})(Q_i-\overline{Q})}} {\sqrt{\sum\limits_{i=1}^k{(S_i-\overline{S})^2}}\sqrt{\sum\limits_{i=1}^k{(Q_i-\overline{Q})^2}}}\texttt{,}
	\label{eq:correlation_pearson}
\end{equation}
where $\overline{S}$ and $\overline{Q}$ denote the mean values of $S$ and $Q$, respectively. To complement the feature ranking and selection step, a PCA is employed to identify features that account for the highest variance in the dataset~\citep{baccar2017identification, jain1999data, jolliffe2002principal, liu1998feature, mathai2022multivariate, sarker2021machine}. The process involves determining the eigenvalues and eigenvectors of the covariance matrix to project the original dataset onto orthogonal Principal Component (PC) subspaces. Each PC is a normalised eigenvector that represents a direction of maximum variance.

Here, the energy indicators, referred to as \textit{x\textsuperscript{(1)}}, are the input for PCA. They are organised into the structure of a feature matrix as follows:
\begin{equation}
	\mathbf{x^{(1)}} = \begin{bmatrix}
	\rax{x_1} \\
	\vdots \\
	\rax{x_m}
	\end{bmatrix}^T = \begin{bmatrix}
	{x}_{1,1} & \cdots & {x}_{1,m} \\
	\vdots  & \ddots & \vdots \\
	{x}_{k,1} & \cdots & {x}_{i,m} 
	\end{bmatrix}\texttt{,}
\end{equation}
with $m$ representing the number of energy indicators and $k$ denoting the number of observations. Accordingly, the PCA transformation is expressed in its general form as~\citep{mathai2022multivariate, macgregor1995statistical}
\begin{equation}
	\mathbf{x^{(1)}} = \mathbf{T}\mathbf{P}^T\texttt{,} 
\end{equation}
where \(\mathbf{P}\) is the loading matrix and \(\mathbf{T}\) is the score matrix. In this context, the loadings represent the coefficients that indicate the degree to which each energy indicator contributes to a principal component. The scores represent the transformed values of the observations, indicating their position in a reduced feature space where a smaller set of principal components captures the main patterns of the original energy indicators. The first principal component captures the largest variance in the energy indicators, while subsequent components account for the remaining variance in decreasing order of magnitude.

To quantify the influence of each feature in the estimation process, an importance score $w^{(m)}$ was calculated for every indicator \(m\). This score is defined as the sum of principal component loadings \(p_i^{(m)}\)and scores \(t_i^{(m)}\) across the total number of components \(N_{c}\):
\begin{equation}
	w^{(m)} = \sum_{i=1}^{N_c}{t_i^{(m)} p_i^{(m)}}\texttt{.}
	\label{eq:importance_score}
\end{equation}  
While correlation analysis and PCA effectively assess feature relevance~\citep{blum1997selection}, real-world sensor data often contains noise, requiring robustness tests to ensure accurate signal representation. To address this, white Gaussian noise was added at an intensity of 5\%, following \citet{ghajari2013identification}. The goal is to identify robust features within an informative and independent subset of descriptors, even in noisy conditions. 

The influence of noise on each candidate feature was quantified using a robustness score $r^{(m)}$ based on the Root-Mean-Square Error (RMSE):
\begin{equation}
	r^{(m)} = 1- \varepsilon_\text{RMSE} \quad\text{and}\quad \varepsilon_\text{RMSE} = \sqrt{\frac{1}{k}\sum\limits_{i=1}^k{(y_i-\hat{y_i})^2}}\texttt{,}
	\label{eq:RMSE}
\end{equation}
where $k$ represents the number of observations, $y$ is the original data, and $\hat{y}$ is the noisy data. Lower RMSE values indicate higher robustness, resulting in higher scores.

To guide feature selection, the robustness score $r^{(m)}$ is combined with the PCA importance score $w^{(m)}$, defining the selection score $s^{(m)}$:  
\begin{equation}
	s^{(m)} = w^{(m)} r^{(m)}\texttt{.}
	\label{eq:selection_score}
\end{equation}

This composite score introduces a unified criterion for ranking features. It ensures that the retained descriptors are relevant, reliable under uncertainties in measurement, and independent by keeping only the strongest representative in each correlation cluster. 

\subsection{Input space validation}
\label{validation_method}

\begin{figure}[htbp]
    \centering
    \begin{subfigure}[t]{\textwidth}
    	\centering
	    \includegraphics[width=\linewidth]{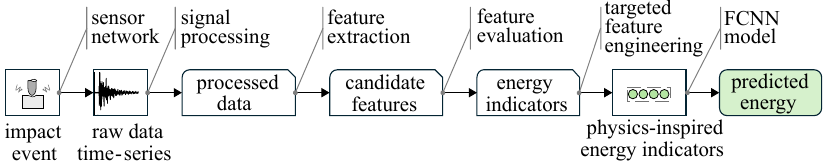}
	    \caption{Model 1: Physics-inspired energy indicators. Descriptors are filtered, ranked, and selected before FCNN estimation.}
	    \label{fig:validation_model_1}
    \end{subfigure}
    \hfill
    \begin{subfigure}[t]{\textwidth}
    	\centering
    	\includegraphics[width=\linewidth]{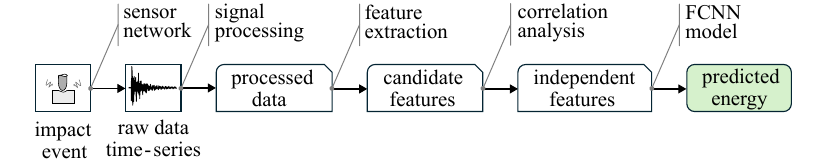}
		\caption{Model 2: Independent features. Correlation analysis removes redundancy before FCNN estimation.}
		\label{fig:validation_model_2}
    \end{subfigure}
    \hfill
    \begin{subfigure}[t]{\textwidth}
    	\centering
    	\includegraphics[width=\linewidth]{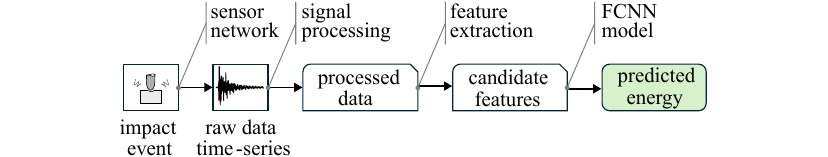}
		\caption{Model 3: Candidate features. All extracted descriptors are input directly to the FCNN.}
		\label{fig:validation_model_3}
    \end{subfigure}
    \hfill
    \begin{subfigure}[t]{\textwidth}
    	\centering
    	\includegraphics[width=\linewidth]{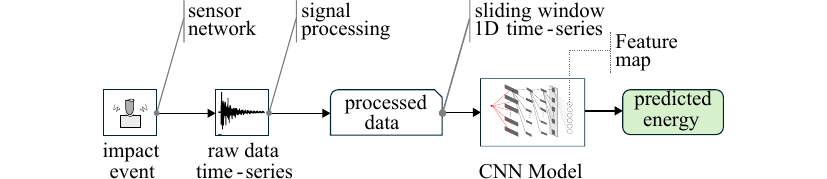}
		\caption{Model 4: CNN. Convolutional layers learn feature maps directly from processed signals.}
		\label{fig:validation_model_4}
    \end{subfigure}
	\caption{High-level overview of the four alternative model architectures considered for impact energy estimation.}
	\label{fig:methods_validation}
\end{figure}

This work employs a machine learning strategy to validate whether the physics-inspired energy indicators enable accurate prediction of impact energy, particularly under limited data conditions. Beyond predictive accuracy, the aim is to examine how models behave when different amounts of domain knowledge are built into the input space. To this end, four configurations are compared: the proposed physics-inspired energy indicators and three reference strategies commonly employed in time-series prediction. These references are an independent feature model that applies correlation analysis, a candidate feature model that retains all extracted descriptors, and a CNN model that learns features directly from processed signals. A high-level overview of the four models is shown in \Cref{fig:methods_validation}.

The models are ordered by decreasing reliance on domain knowledge. The proposed physics-inspired energy indicators model (Model 1, \Cref{fig:validation_model_1}) incorporates the highest level of knowledge through targeted feature engineering based on statistical and prior knowledge criteria. The independent feature model (Model 2, \Cref{fig:validation_model_2}) reduces this knowledge by skipping targeted ranking and selection, retaining only a correlation filter to remove redundancy; such filtering is well established in data science for mitigating overfitting~\citep{muir2021damage, daghigh2024explainable, hall1999correlation}. The candidate feature model (Model 3, \Cref{fig:validation_model_3}) further relaxes prior knowledge by retaining the complete set of extracted descriptors without any filtering or refinement, reflecting approaches commonly adopted in structural diagnostics~\citep{sharif2018impact, oskouei2009wavelet}.

At the lowest level of embedded knowledge, the CNN model (Model 4, \Cref{fig:validation_model_4}) represents a purely data-driven baseline. Here, convolutional layers learn feature maps directly from segmented time-series signals, with no manual feature design. This end-to-end approach is widely adopted in SHM for tasks that rely on automatic feature discovery~\citep{lecun1998gradient, tabian2019convolutional, ai2024acoustic}. Taken together, this comparative analysis assesses whether embedding domain knowledge yields measurable benefits in predictive performance under data constraints.

\section{Datasets}
\label{system_description}

\begin{figure}[htbp]
	\centering
	\begin{subfigure}[t]{2.51cm}
	    \includegraphics[width=\linewidth]{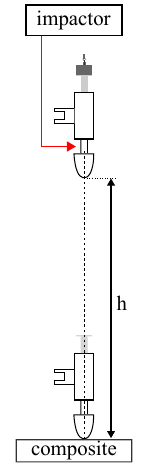}
	    \caption{\centering}
	    \label{fig:dt_assembly}
	\end{subfigure}
	\qquad
	\begin{subfigure}[t]{8cm}
	    \includegraphics[width=\linewidth]{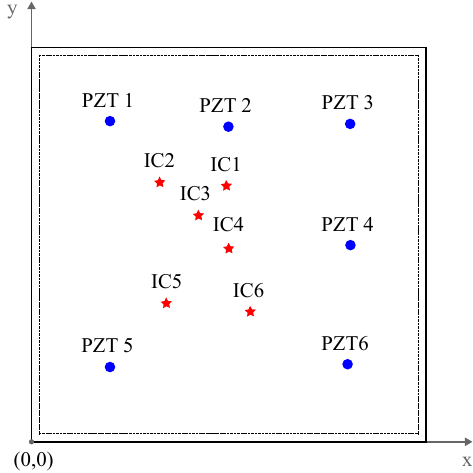}
	    \caption{\centering}
	    \label{fig:experimentaldesign_coupon}
	\end{subfigure}
	\caption{Schematic impact test configuration; \textcolor{blue}{$\bullet$} piezoelectric sensors (PZT) and \textcolor{red}{$\star$} Impact Locations (IC): (a) drop-tower assembly with impact height ($h$) set by impact energy; (b) square composite panel ($1000 \times  1000 \times 3.55$~mm). Adapted from \citet{marinho2025evaluating}.}
	\label{fig:subfig_setup}
\end{figure}

To validate the proposed input space selection method, both synthetic and experimental datasets were generated from intermediate mass impact tests~\citep{olsson2000mass}. All datasets share the same structural configuration: a square composite panel with nominal dimensions of 1000~mm $\times$ 1000~mm $\times$ 3.55~mm. \Cref{fig:subfig_setup} illustrates the impact test setup, and \Cref{tab:setup_coordinates} lists the sensors and impact location coordinates.
\begin{table}[htbp]
	\centering
	\caption{Impact locations and sensor network coordinates.}
	\label{tab:setup_coordinates}
	\begin{tabular}{@{}lcc@{}}
		\toprule
		\multirow{2}[0]{*}{\textbf{ID}} & \multicolumn{2}{c}{\textbf{Coordinates [mm]}} \\
		\cmidrule(lr){2-3}
		& \textit{x} & \textit{y} \\   
		\midrule
		PZT1  & 188   & 786 \\
		PZT2  & 478   & 780 \\
		PZT3  & 779   & 781 \\
		PZT4  & 779   & 481 \\
		PZT5  & 188   & 172 \\
		PZT6  & 768   & 174 \\
		IC1   & 477   & 600 \\
		IC2   & 328   & 629 \\
		IC3   & 402   & 555 \\
		IC4   & 479   & 480 \\
		IC5   & 349   & 348 \\
		IC6   & 574   & 282 \\
		\bottomrule
	\end{tabular}
\end{table}

The numerical dataset was used exclusively for feature evaluation and sensitivity analysis under controlled, noise-free conditions, as described previously in \Cref{subsec:ANOVA_methods}. It was derived from an explicit finite element model validated against experimental findings in \citet{bezes2024validation}. The simulations replicate the panel geometry, boundary conditions, and material properties while spanning a representative range of impact energies and locations. This dataset corresponds to the confirmation experiments introduced in \Cref{tab:testmatrix_conf} for the ANOVA evaluations. Each simulated waveform was stored as a time series, and the statistical values are summarised in \Cref{tab:expstats}.

The experimental dataset was used for feature ranking and selection (see \Cref{subsec:target_engineering}) and model validation (see \Cref{validation_method}). A total of 66 waveforms were acquired from impact tests conducted in accordance with \citet{ASTMD7136}, covering impact energies ranging from 3.81~J to 85.37~J across both pristine and damaged states. Impacts were applied at multiple panel locations (IC1-IC6 in \Cref{fig:subfig_setup}), using three different impactor diameters (16~mm, 25~mm, and 50~mm) and three impactor masses (0.776~kg, 1.154~kg, and 2.356~kg). The signals were recorded using a six-channel surface-mounted piezoelectric sensor network and stored as univariate time series. Further details regarding the material system, instrumentation, and acquisition procedures are provided in \citet{marinho2025evaluating}.

To further test robustness, additional variability was introduced by adding 5\% Gaussian noise to the experimental signals. This augmentation simulates harsher acquisition conditions while preserving the original measurements, yielding a total of 132 samples for model development. Both the original and noise-augmented experimental signals were used in the analysis. A summary of statistics of impact energies is provided in \Cref{tab:numstats}. The dataset is intentionally unbalanced, sparse, and limited in scope to reflect realistic constraints in SHM applications. Notably, the median impact energy (7.94~J) is substantially lower than the maximum recorded value (85.37~J), indicating a skewed distribution with a greater density of low-energy events.
\begin{table}[htbp]
	\centering
    \caption{Impact energy statistics.}
    \label{tab:energy_stats}
  	\begin{subtable}[t]{0.47\textwidth}
	  	\centering
	  	\caption{Numerical dataset.}
	  	\label{tab:expstats}
	    \begin{tabular}{@{}lc@{}}
		    \toprule
		    Number of samples         & 8 \\
		    Number of unique targets  & 2 \\
		    Minimum energy [J]        & 2.00 \\
		    Maximum energy [J]        & 20.00 \\
		    Mean energy [J]           & 11.00 \\
		    Median energy [J]         & 11.00 \\
		    Standard deviation [J]    & 9.49 \\
		    \bottomrule
	    \end{tabular}
    \end{subtable}
 	\qquad
	\begin{subtable}[t]{0.47\textwidth}
		\centering
		\caption{Experimental dataset.}
		\label{tab:numstats}
	    \begin{tabular}{@{}lc@{}}
		    \toprule
		    Number of samples         & 132 \\
		    Number of unique targets  & 66 \\
		    Minimum energy [J]        & 3.81 \\
		    Maximum energy [J]        & 85.37 \\
		    Mean energy [J]           & 23.54 \\
		    Median energy [J]         & 7.94 \\
		    Standard deviation [J]    & 20.26 \\
		    \bottomrule
	    \end{tabular}
    \end{subtable}
\end{table}

\section{Results and discussion}
\label{sec:results}

The results obtained from the proposed methodology for developing and validating an optimised input space for estimating impact energy are presented herein. The multi-domain candidate features are tested for four key properties: energy sensitivity, linear independence, relevance, and robustness to noise.

\subsection{Feature evaluation}
\label{ANOVA}

Feature evaluation was used to identify which candidate features, defined in \Cref{subsec:energy_indicators}, respond significantly to variations in impact energy. To this end, an F-test was applied as the statistical test of the null hypothesis. The null hypothesis $H_0$ states that the mean values of a feature remain equal across all impact energy levels, meaning the feature is not sensitive to energy variation. Rejecting $H_0$ ($H_0 = \text{False}$) shows that at least one group mean differs, and the feature is therefore classified as energy-sensitive. The results of this evaluation are presented in \Cref{tab:feature_evaluation}, where energy-sensitive features are highlighted in bold text.
\begin{table}[htbp]
	\centering
    \caption{Feature evaluation with $F_{crit} = 7.7$.}
    \label{tab:feature_evaluation}
    \begin{tabular}{@{}llcll@{}}
        \toprule
        \textbf{Domain} & \textbf{ID} & \textit{\textbf{F$_{calc}$}} & \textit{\textbf{H$_0$}} & \textbf{Energy-sensitive?} \\
        \midrule
        \textbf{Time} 				& \textbf{RMS} 	& \textbf{43.2} 		& \textbf{False} 	& \textbf{Yes} \\
        							& \textbf{TE} 	& \textbf{16.6} 		& \textbf{False} 	& \textbf{Yes} \\
        							& \textbf{PA} 	& \textbf{51.2} 		& \textbf{False} 	& \textbf{Yes} \\
        							& \textbf{EPR} 	& \textbf{32.7} 		& \textbf{False} 	& \textbf{Yes} \\
        							& \textbf{RA} 	& \textbf{30.6} 		& \textbf{False} 	& \textbf{Yes} \\
        							& CTP 	& 7.0 		& True 		& No \\
        							& RT 	& 2.2 		& True 		& No \\
        							& NDA 	& 1.8 		& True 		& No \\
        \textbf{Frequency} 		& \textbf{PCR} 	& \textbf{354.2} 	& \textbf{False} 	& \textbf{Yes} \\
        							& ROFF 	& 0.9 		& True 		& No \\
									& CF 	& 1.3 		& True 		& No \\
									& \textbf{WPF} 	& \textbf{9.0} 		& \textbf{False} 	& \textbf{Yes} \\
									& \textbf{PF} 	& \textbf{9.0} 		& \textbf{False} 	& \textbf{Yes} \\
									& RON 	& 1.0 		& True 		& No \\
		\textbf{Time-Frequency} 	& \textbf{AME} 	& \textbf{17.2} 		& \textbf{False} 	& \textbf{Yes} \\
									& \textbf{AM} 	& \textbf{62.6} 		& \textbf{False} 	& \textbf{Yes} \\
									& \textbf{DM} 	& \textbf{11.0} 		& \textbf{False} 	& \textbf{Yes} \\
									& DME 	& 3.6 		& True 		& No \\
        \bottomrule
    \end{tabular}
\end{table}

In the time domain, Root Mean Square (RMS), Peak Amplitude (PA), Transmitted Energy (TE), Energy Peak Ratio (EPR), and Rise Angle (RA) showed strong sensitivity to impact energy. These parameters are intrinsically linked to signal amplitude and energy content, which consistently scale with impact energy. Specifically, RMS and PA quantify signal power and maximum amplitude, respectively, while TE measures the total transmitted energy. EPR and RA capture transient signal characteristics, including rise time and energy distribution, which vary with impact severity~\citep{jang2015impact, ghajari2013identification, tabian2019convolutional, guel2020data}.

In the frequency domain, Peak Centroid Ratio (PCR), Peak Frequency (PF), and Weighted Peak Frequency (WPF) responded to spectral shifts induced by varying impact energy: higher energy produced shorter impacts and broader spectra, while lower energy produced longer impacts and lower-frequency spectra. Mechanical models confirm that impact duration and energy level govern the force profile and spectral content~\citep{WITKOWSKI2022108417}. These shifts reflect modal participation and redistribution of vibrational energy consistent with frequency-dependent structural behaviour~\citep{Stull_2011, vsofer2022acoustic, marinho2025evaluating, marafini2024investigation}.

Among the time–frequency features, AM, AME, and DM were sensitive to impact energy, capturing propagation and attenuation mechanisms critical to energy partitioning~\citep{oskouei2009wavelet, fotouhi2015investigation}. AM and AME are associated with low-frequency stress wave propagation and scale with energy input, whereas DM captures abrupt high-frequency changes, reflecting localised phenomena such as impact events~\citep{mallat1989theory}.

In contrast, several features showed limited sensitivity to impact energy due to their underlying definitions and dependencies. Non-Dimensional Amplitude (NDA) cancels amplitude effects by normalisation, while Detailed Max Energy (DME) is dominated by stochastic high-frequency content such as noise and scattering, weakening its link to input energy~\citep{roberts2013computational, oskouei2009wavelet, guel2020data}. Counts to Peak (CTP) and Rise Time (RT) depend on wave speed and geometry, which are independent of energy level~\citep{farrar2012structural}. Centroid Frequency (CF) and Roll-OFF/ON frequencies (ROFF, RON) describe relative spectral distribution rather than absolute amplitude, thus reflecting structural characteristics rather than impact energy~\citep{farrar2012structural}.

In summary, the results confirm theoretical expectations from structural dynamics: features sensitive to amplitude and transient signal behaviour serve as effective indicators of impact energy~\citep{lu2003energy, melis2018dynamic, shafieerobustness}. The F-test thus ensures that only statistically relevant features are retained for subsequent stages of analysis. Based on the evaluation presented in \Cref{tab:feature_evaluation}, the selected features for further analysis include RMS, TE, PA, EPR, RA, PCR, WPF, PF, AME, AM, and DM.

\subsection{Feature ranking and selection} \label{sec:ranking_selection}

The feature ranking and selection are based on the energy-sensitive features identified by ANOVA (see \Cref{ANOVA}). To ensure consistent comparison, each feature was normalised using min-max scaling~\citep{patro2015normalization}, preserving underlying trends and eliminating amplitude-dependent biases in ranking metrics.

\begin{figure}[htbp]
	\centering
	\includegraphics[width=13.21cm]{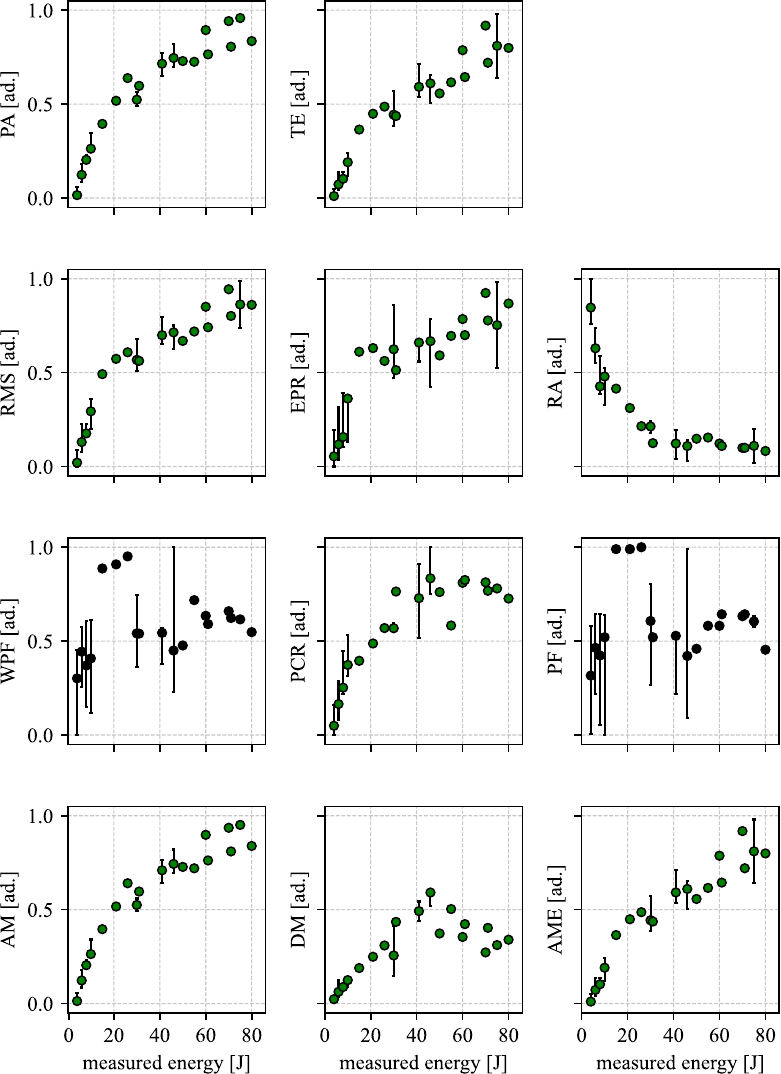}
	\caption{Univariate range plots of energy indicators for EDA. Features that exhibit either monotonicity or trendability are highlighted in green.}
	\label{fig:EDA_boxplot}
\end{figure}

As an initial step, Exploratory Data Analysis (EDA) was employed to assess whether features behave consistently with respect to impact energy. Two properties guided this qualitative assessment: monotonicity and trendability. Monotonicity denotes the degree to which a feature changes in a single direction without reversals, while trendability denotes the consistency of this curve shape across different energy levels. Both properties are widely used to evaluate whether a feature provides a reliable measure of system response~\citep{6544227,6824783}. The range plots in \Cref{fig:EDA_boxplot} show the relationships between normalised features and impact energy, with medians marked and whiskers extending to the most extreme data points that fall within 1.5 times the interquartile range~\citep{NUZZO2016268}. They capture both the spread of the data at each level and the shift of the central value, which are key to evaluating feature consistency. Features that demonstrate either monotonicity or trendability are highlighted in green, emphasising their underlying trends.

Amplitude-based descriptors, including Peak Amplitude (PA), Transmitted Energy (TE), Root Mean Square (RMS), Approximation Max (AM), and Approximation Max Energy (AME), demonstrate a consistent increase with impact energy. Initially, linear trends are observed at lower energy levels; however, as energy increases, non-linear growth becomes evident. This shift indicates potential geometric or material non-linearities that may arise from wave–structure interactions or damage mechanisms, as highlighted by \citet{melis2018dynamic}. These non-linear effects may not only cause features to saturate but can also lead to accelerated growth, deviating from proportional scaling with energy input. The Energy Peak Ratio (EPR) and Peak Centroid Ratio (PCR) increase similarly with impact energy and exhibit a clear trend. However, EPR loses monotonicity at the highest levels, where saturation effects emerge. The rise angle (RA) shows an inverse yet monotonic relationship with impact energy, reflecting transient response dynamics and enhanced damping effects, as demonstrated in previous studies~\citep{anderson2012basic,hassani2021structural}.

In the frequency domain, the Peak Frequency (PF) and Weighted Peak Frequency (WPF) exhibit considerable scatter and irregular trends with respect to impact energy. Although sensitive to excitation changes, their consistency is affected by abrupt shifts in dominant modes and spectral redistribution across energy levels~\citep{kullaa2011distinguishing}. These factors introduce heteroscedasticity, reducing their suitability for stand-alone energy estimation. Despite these limitations, frequency-based features offer valuable supplementary insight into structural dynamics and enhance the overall characterisation of impact response.

Furthermore, the Detailed Max (DM) shows a piecewise linear pattern, increasing with energy up to about 45~J and then declining. Although this indicates a clear trend, the bi-linear behaviour lacks monotonicity because it reverses direction at higher levels. The lack of monotonicity may limit its suitability for energy estimation, as a single DM value can correspond to multiple energy levels, creating ambiguity and reducing predictive accuracy. The decline at higher energies suggests a shift from elastic wave propagation to attenuation of high-frequency components, possibly driven by geometric non-linearities or the onset of damage. Literature supports this interpretation, reporting that structural defects scatter stress waves and dissipate vibrational energy into heat at elevated excitation levels~\citep{muir2021damage, fotouhi2015investigation, chen2014damage}.

\begin{figure}[htbp]
	\centering
	\begin{subfigure}[t]{0.48\textwidth}
	    \centering
	    \includegraphics[width=\linewidth]{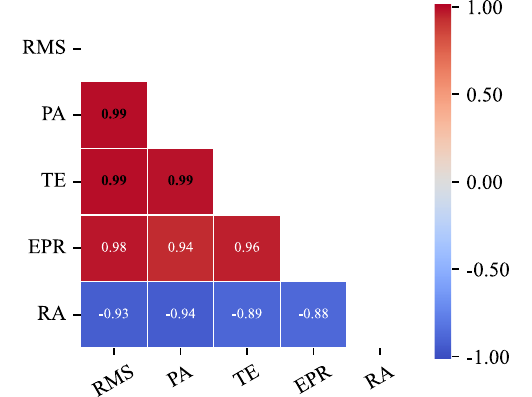}
	    \caption{\centering Time-domain.}
	    \label{fig:correlation_analysis_time_domain}
	\end{subfigure}
	\hfill
	\begin{subfigure}[t]{0.48\textwidth}
	    \centering
	    \includegraphics[width=\linewidth]{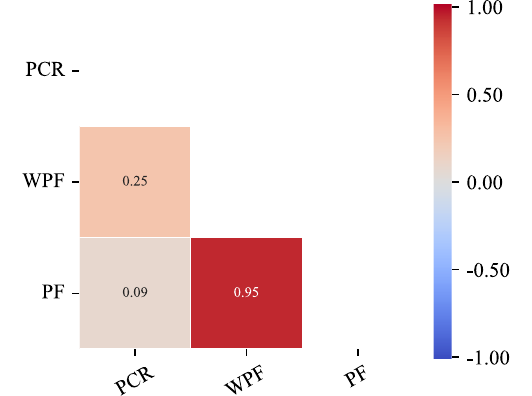}
	    \caption{\centering Frequency-domain.}
	    \label{fig:correlation_analysis_frequency_domain}
	\end{subfigure}
	\hfill
	\begin{subfigure}[t]{0.48\textwidth}
	    \centering
	    \includegraphics[width=\linewidth]{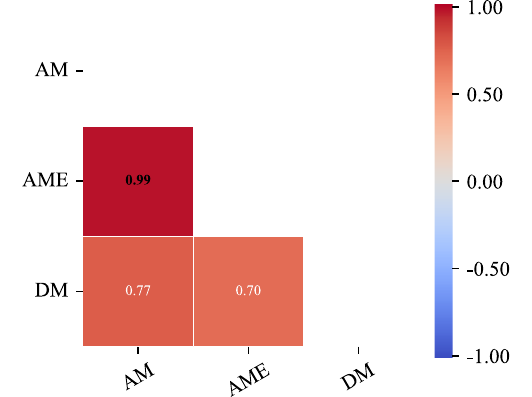}
	    \caption{\centering Time-frequency domain.}
	    \label{fig:correlation_analysis_time_frequency_domain}
	\end{subfigure}
	\caption{Correlation analysis results.}
	\label{fig:correlation_analysis}
\end{figure}

To complement the univariate analysis, correlation analysis was performed to assess redundancy and complementarity among descriptors across signal domains. This step mitigates information overlap and supports input space disentanglement, which is essential for effective feature representations~\citep{saeedifar2020damage}. Groups in \Cref{tab:feature_ranking_coupon} are coded by domain and statistical similarity: T for time-domain features, F for frequency-domain features, and W for time–frequency (wavelet-based) features. Within each domain, the numbering (e.g., T1, T2) identifies correlated clusters derived from Pearson correlation coefficients (\Cref{eq:correlation_pearson}). These relationships are visualised in \Cref{fig:correlation_analysis}, highlighting domain-specific associations. The additional columns in \Cref{tab:feature_ranking_coupon} are introduced here for completeness but will be explained in detail after the PCA and robustness analyses later in this section.
\begin{table}[htbp]
	\centering
	\begin{threeparttable}
	    \caption{Feature ranking and selection at coupon level.}
	    \label{tab:feature_ranking_coupon}
		\begin{tabular}{@{}lllccccc@{}}
			\toprule
			\textbf{Domain} & \textbf{Group} & \textbf{ID} & $\mathbf{w^{(m)}}$ & $\mathbf{r^{(m)}}$ & $\mathbf{s^{(m)}}$ & \textbf{Rank} & \textbf{Evaluation} \\
			\midrule
			\textbf{Time} 				& T1 & RMS 	& 0.80 & 0.98 & 0.78 & 2 & $\ast$ / $\bullet$ \\
										& T1 & TE  	& 0.80 & 0.97 & 0.77 & 3 & $\ast$ / $\bullet$ \\
										& T1 & PA  	& 0.80 & 0.99 & 0.79 & 1 & $\ast$ / $\bullet\bullet$ \\
										& T2 & EPR 	& 0.76 & 0.95 & 0.73 & 1 & $\ast\ast$ / $\bullet\bullet$ \\
										& T3 & RA  	& 0.76 & 0.92 & 0.70 & 1 & $\ast\ast$ / $\bullet\bullet$ \\
			\textbf{Frequency} 			& F1 & PCR 	& 0.78 & 0.89 & 0.69 & 1 & $\ast\ast$ / $\bullet\bullet$ \\
										& F2 & WPF 	& 0.60 & 0.91 & 0.55 & 1 & $\ast\ast$ / $\bullet\bullet$ \\
										& F3 & PF 	& 0.45 & 0.93 & 0.42 & 1 & $\ast\ast$ / $\bullet\bullet$ \\
			\textbf{Time-Frequency} 	& W1 & AME 	& 0.80 & 0.99 & 0.79 & 1 & $\ast$ / $\bullet\bullet$ \\
										& W1 & AM  	& 0.80 & 0.98 & 0.79 & 2 & $\ast$ / $\bullet$ \\
										& W2 & DM  	& 0.72 & 0.63 & 0.46 & 1 & $\ast\ast$ / $\diamond$ \\
			\bottomrule
		\end{tabular}
		\begin{tablenotes}[flushleft]
		    \footnotesize
		    \item \textbf{Legend:}
		    \item \textbf{Scores:} $\mathbf{w^{(m)}}$ importance score (\Cref{eq:importance_score}); $\mathbf{r^{(m)}}$ robustness score (\Cref{eq:RMSE}); $\mathbf{s^{(m)}}$ selection score (\Cref{eq:selection_score}).
		    \item \textbf{Evaluation:} $\diamond$ not stable; $\ast$ not independent; $\ast\ast$ independent; $\bullet$ relevant; $\bullet\bullet$ most relevant.
		    \item \textbf{Rank:} Ranking per group based on selection score.
		\end{tablenotes}
	\end{threeparttable}
\end{table}

In the time-domain (\Cref{fig:correlation_analysis_time_domain}), amplitude-based features (PA, TE, RMS) exhibit strong intercorrelation, driven by their shared sensitivity to signal magnitude and energy content. These descriptors form a coherent cluster, demonstrating predictable scaling with wave amplitude in the elastic regime. By contrast, RA exhibits a high negative correlation with these features, consistent with its inverse relationship with energy identified during EDA.

In the frequency-domain (\Cref{fig:correlation_analysis_frequency_domain}), PF and WPF showed a strong correlation ($p = 0.95$), reflecting mutual sensitivity to dominant spectral content. However, PCR exhibited weak correlations with both PF ($p = 0.09$) and WPF ($p = 0.25$), suggesting that it represents spectral shape rather than frequency magnitude. This independence makes PCR complementary to PF and WPF, since together they provide distinct yet related information on modal behaviour, as also noted in studies on spectral clustering~\citep{alamdari2017spectral}.

In the time-frequency domain (\Cref{fig:correlation_analysis_time_frequency_domain}), AM and AME exhibit a near-perfect correlation ($p = 0.99$), reflecting shared sensitivity to low-frequency stress waves and energy distribution. DM shows a moderate correlation with both AM and AME ($p = 0.77$ and $p = 0.70$, respectively), indicating its complementary role in capturing high-frequency transients while maintaining sensitivity to energy variation.

\begin{figure}[htbp]
	\centering
	\includegraphics{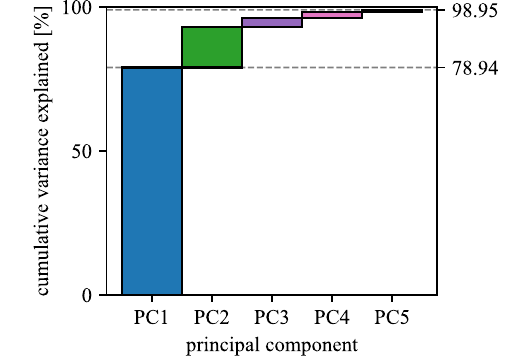}
	\caption{Explained variance of principal components.}
	\label{fig:PCA_explained_variance}
\end{figure}

A Principal Component Analysis (PCA) was conducted in parallel with correlation analysis to evaluate the variance structure of the feature set. In accordance with previous literature~\citep{sibil2012optimization}, the number of retained components was defined to capture over 95~\% of the total variance. As illustrated in \Cref{fig:PCA_explained_variance}, the first five components collectively account for 98.95~\% of the cumulative variance, with PC1 alone explaining 78.94~\%.

The PCA-derived importance scores, summarised under the column $\mathbf{w^{(m)}}$ (\Cref{eq:importance_score}) in \Cref{tab:feature_ranking_coupon}, confirm the trends identified in the exploratory and correlation analyses. RMS, TE, and PA, all from the time domain, each attained a score of 0.80, highlighting their dominant contribution to the variance structure. Their similar scores reflect the strong correlations observed earlier, illustrating how PCA assigns comparable importance to highly correlated features. EPR and RA both scored 0.76; in particular, the RA contribution reflects its distinct variance pattern, which aligns with its inverse relationship to impact energy, as also indicated by the negative correlation coefficients in \Cref{fig:correlation_analysis_time_domain}.

In the frequency domain, PCR achieved the highest score (0.78), confirming its statistical independence from PF and WPF, as indicated by weak correlations. The lower scores for WPF (0.60) and PF (0.45) correspond to the irregular, dispersed trends observed in the exploratory analysis. In the time-frequency domain, AM and AME both achieved scores of 0.80, consistent with their near-perfect correlation. DM, with a moderate score of 0.72, complements this pair by capturing additional high-frequency dynamics not fully represented by AM or AME.

\begin{figure}[htbp]
	\centering
	\includegraphics[width=14cm]{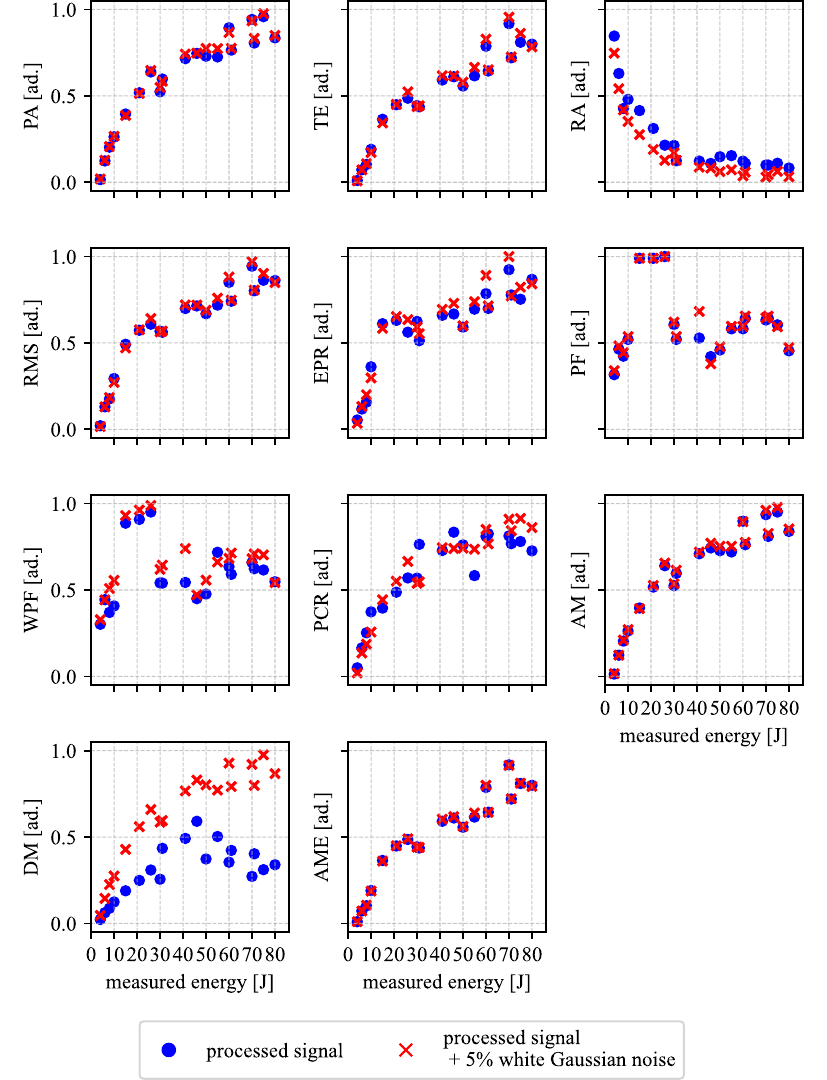}
	\caption{Robustness of features under added noise.}
	\label{fig:stability_test}
\end{figure}

A robustness assessment complemented the feature ranking and selection step with a noise-sensitivity analysis, in which 5\% Gaussian noise was added to the dataset.  Robustness scores $r^{(m)}$ (\Cref{eq:RMSE}), reported in \Cref{tab:feature_ranking_coupon}, quantified the resilience of each feature against these perturbations. \Cref{fig:stability_test} illustrates the resultant effects, presenting median values for each energy level to elucidate underlying trends.

Amplitude-based features (PA, RMS, TE, AM, and AME) demonstrated high robustness ($r^{(m)} \geq 0.97$), maintaining consistent energy trends in the presence of noise. This robustness is consistent with their strong energy sensitivity and dominant contribution to variance. As previously confirmed by both exploratory and PCA analyses, these descriptors showed stable monotonic behaviour and high importance scores. Furthermore, since additive white Gaussian noise affects all dimensions uniformly in PCA~\citep{pyatykh2012image}, these high-importance features tend to preserve their relative relationships under perturbation. This robustness results from their signal variance substantially exceeding the noise level, whereas features with lower importance scores are more vulnerable to saturation and distortion effects~\citep{hong2018asymptotic, hong2021heppcat}.

The remaining time-domain features, EPR and RA, also exhibited high robustness ($r^{(m)} \geq 0.92$), confirming their reliability for impact identification tasks. In contrast, frequency-domain features demonstrated moderate robustness, with PCR, WPF, and PF achieving scores ranging from 0.89 to 0.93. Although these features remained sufficiently robust, they were more affected by noise and modal shifts than amplitude-based descriptors.

Within the time-frequency domain, DM was significantly more sensitive to noise ($r^{(m)} = 0.63$), as indicated by increased scatter under perturbation. This greater sensitivity arises because noise predominantly affects high-frequency components, which are captured by the detailed coefficients~\citep{basu2005identification}, limiting its applicability under realistic operating conditions.

The selection score $s^{(m)}$ (\Cref{eq:selection_score}), comprising both importance and robustness scores, is summarised in \Cref{tab:feature_ranking_coupon} and directs the final evaluation of descriptors. This phase synthesises findings from the correlation structure, PCA-based relevance, and noise-sensitivity analyses. Rather than applying a strict cut-off, the mathod guides informed feature selection. Appropriate descriptors strengthen confidence in the input space, while weaker ones expose its limitations and indicate where predictive accuracy may be improved. Finally, the selection score ranks stable descriptors of comparable importance within correlated clusters, enabling a consistent choice among redundant features.

In the time domain, all energy-sensitive features exhibited sufficient robustness. However, correlation analysis revealed redundancy among RMS, TE, and PA (group T1). PA was retained due to its highest selection score in this cluster. EPR and RA also demonstrated robustness and independence and were therefore included in the final feature set. In the frequency domain, PCR, WPF, and PF fulfilled the criteria for independence and robustness and were retained accordingly. Within the time-frequency domain, AM and AME displayed a strong correlation, with AME selected due to its slightly superior robustness. DM, although energy-sensitive, failed to meet the robustness criterion and was excluded from the final input layer.

In conclusion, feature selection was accomplished through a process of elimination, refining the input space to include only robust, independent, and energy-sensitive descriptors. The final set of energy indicators for impact energy estimation comprises PA, EPR, RA, PCR, WPF, PF, and AME. By integrating descriptors from the time, frequency, and time–frequency domains, the selection process covers the mechanisms identified during extraction, yielding an input space guided by domain knowledge.

\subsection{Input space validation}

The constructed input space, based on selected physics-inspired energy indicators, was validated for its effectiveness in predicting impact energy. Its performance was compared against three reference methodologies, as outlined in \Cref{validation_method}. For clarity, the input space of each model is summarised below:
\begin{itemize}
	\item \textbf{Model 1} (Physics-inspired energy indicators): PA, EPR, RA, PCR, WPF, PF, and AME, as defined by the proposed feature selection process;
	\item \textbf{Model 2} (Independent features): PA, RT, CTP, RA, EPR, NDA, CF, PF, WPF, PCR, RON, ROFF, DM, and DME, obtained from Pearson correlation analysis of the full candidate set without distinguishing domains;
	\item \textbf{Model 3} (Candidate features): all descriptors listed in \Cref{tab:feature_extraction};
	\item\textbf{Model 4} (CNN): abstract feature representations through convolutional operations.
\end{itemize}

All models were trained on identical data partitions to ensure a fair comparison. The dataset was split into training (80\%), validation (10\%), and test (10\%) subsets and included both original measurements and augmented data. Model implementation details, architectures, and hyperparameters are provided in the following subsections.

\subsubsection{FCNN implementation}

Artificial Neural Networks (ANNs) were employed to predict energy using the input spaces of Models 1, 2, and 3. Fully Connected Neural Networks (FCNNs) were selected for their effectiveness in capturing non-linear relationships between signal-derived features and impact energy through iterative optimisation~\citep{Zargar_2020,bishop1995neural, haykin1994neural, ghajari2013identification}. 

\begin{figure}[htbp]
	\centering
	\includegraphics{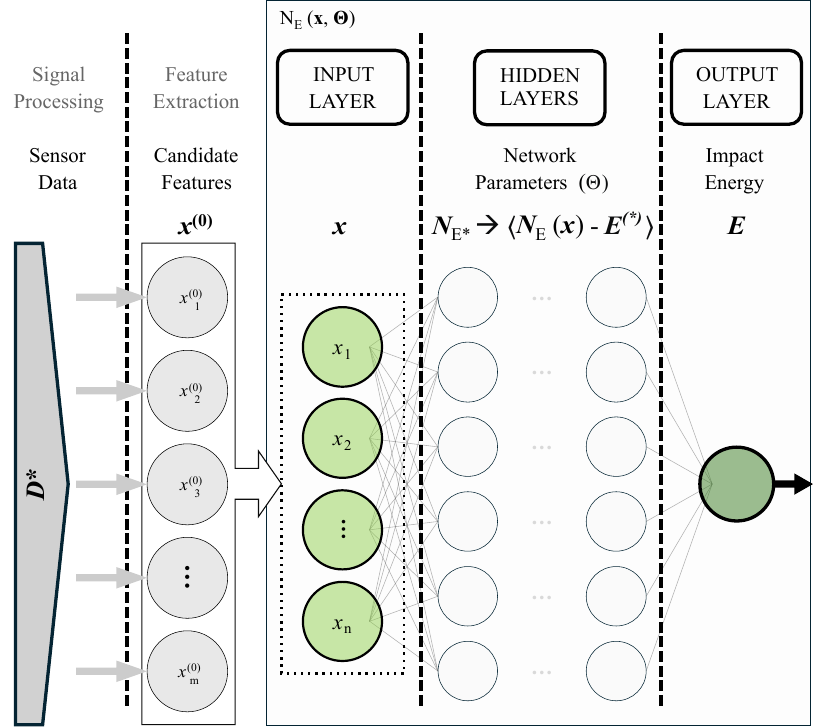}
	\caption{Schematic of the FCNN (Models 1--3).}
	\label{fig:nn_structure}
\end{figure}

The overall workflow is illustrated in \Cref{fig:nn_structure} and consists of four main steps. First, raw data \textbf{\textit{D}}\textsuperscript{*} were pre-processed using the signal processing framework described in \citet{marinho2025evaluating}. Second, candidate features were extracted as outlined in \Cref{sec:selection_method}. Third, the input space was defined, with the input vector differing across models. Fourth, the FCNN used Tanh activation functions~\citep{sharma2017activation} to establish non-linear mappings between the input features and the target impact energy. Training minimised the Mean Squared Error (MSE) between the network output \textbf{\textit{E}} and the experimental ground truth \textbf{\textit{E}}\textsuperscript{(*)}, producing a set of optimised parameters $\mathbf{\Theta}$. Model performance was finally assessed on independent test data using standard error metrics.

The network settings were selected through a grid search over the hyperparameter space summarised in \Cref{tab:grid_search_fnn}. In this process, three hyperparameters were varied in parallel: the size of the fully connected layer ($n_h$), the number of fully connected layers ($L_h$), and the learning rate ($lr$). Each configuration was tested with five-fold cross-validation, ensuring that the chosen parameters were generalised across the dataset~\citep{DIPIETRANGELO2023109621}. For each fold, the dataset was partitioned into independent 80/20 train and test splits, and the mean coefficient of determination ($R^2$) across folds was used as the performance criterion.
\begin{table}[htbp]
	\centering
	\caption{Hyperparameter search space for tuning the FCNN model (Models 1–3).}
	\label{tab:grid_search_fnn}
	\begin{tabular}{@{}lll@{}}
		\toprule
		\textbf{Category} 	& \textbf{Parameter} & \textbf{Values} \\
		\midrule
		Model architecture & fully connected layer size, $n_h$ 			& 32, 64, 128 \\
		                   & number of fully connected layers, $L_h$  	& 2, 3 \\
		\arrayrulecolor{gray!50}		
		\midrule
		Training setup     & learning rate, \textit{lr}     				& 0.01, 0.001 \\
		\arrayrulecolor{black}
		\bottomrule
	\end{tabular}
\end{table}

\begin{figure}[htbp]
    \centering
    \includegraphics{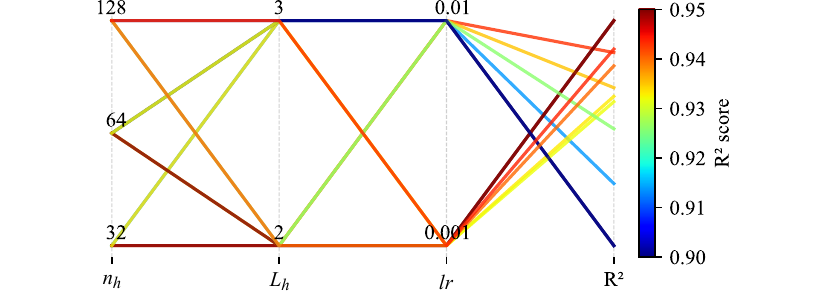}
    \caption{Grid search with $k$-fold cross-validation results. ($n_h$: fully connected layer size; $L_h$: number of fully connected layers; and \textit{lr}: learning rate).}
    \label{fig:grid_search}
\end{figure}

The results of this procedure are shown in \Cref{fig:grid_search}, where the colour scale indicates the average $R^2$ obtained for each parameter combination. The grid search identified the best performance at $n_h = 32$, $L_h = 2$, and $lr = 1 \times 10^{-3}$. The selected configuration, together with all implementation details of the FCNN, is reported in \Cref{tab:fnn_hyperparams}. These settings were used in the residual and performance analyses.
\begin{table}[htbp]
	\centering
	\caption{Implementation details used for impact energy predictions using FCNN (Models 1–3).}
	\label{tab:fnn_hyperparams}
	\begin{tabular}{@{}llc@{}}
		\toprule
		\textbf{Parameter} & \textbf{Value} \\
		\midrule
		\textbf{Network architecture} & \\
		Architecture 							& Fully-Connected Neural Network \\
		Fully connected layers, $n_h$ 			& 2 \\
		Fully connected layer size, $L_h$ 		& 32 \\
		Output layer dimension 				& 1 \\
		Activation function 					& Tanh \\
		Batch normalisation 					& True \\
		\arrayrulecolor{gray!50}
		\midrule
		\textbf{Training procedure} & \\
		Optimizer 								& Adam \\
		Criterion 								& MSELoss \\
		Learning rate, $lr$ 					& $1 \times 10^{-3}$ \\
		Batch size 								& Full-batch \\
		Max epochs 								& 10000 \\
		Patience 								& 1000 \\
		\arrayrulecolor{black}
		\bottomrule
	\end{tabular}
\end{table}

\subsubsection{CNN implementation}

\begin{figure}[htbp]
\centering
    \includegraphics{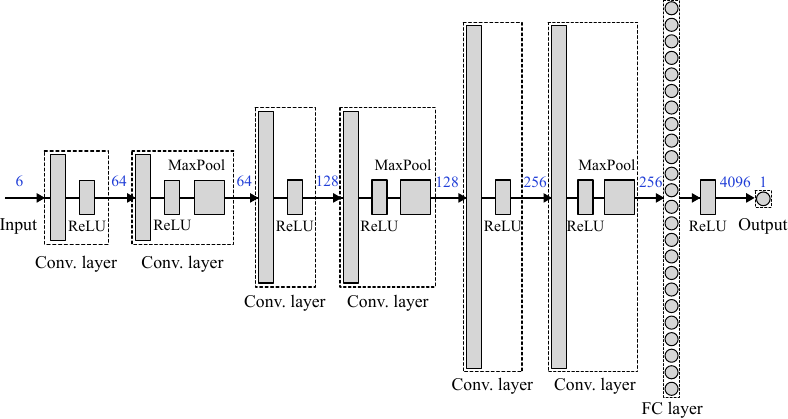}
    \caption{Architecture of the CNN used for impact energy estimation.}
    \label{fig:CNN_architecture}
\end{figure}

The CNN acts as a regression model for predicting impact energy. Unlike previous models that rely on explicit feature selection, the CNN learns feature maps directly from measured impact responses. The network architecture is shown in \Cref{fig:CNN_architecture}. It has six input channels that process 1-D time-series measurements as overlapping time slices, generated with a sliding window of 1000 samples and a shift of 200 samples (80\% overlap). The input layer is followed by convolutional layers with 64, 128, and 256 channels, each combined with ReLU activation and periodic max-pooling. A fully connected layer produces a single output corresponding to the predicted impact energy. During training, the model with the lowest loss per iteration was saved for prediction. A patience threshold of 15 epochs is applied, meaning training stopped early if no further improvement was observed for 15 consecutive iterations. The corresponding hyperparameters and configuration choices are listed in \Cref{tab:cnn_hyperparams}. This configuration was obtained through a focused manual search to optimise windowing and training, which was sufficient for the scope of this study.
\begin{table}[htbp]
	\centering
	\caption{Implementation details used for impact energy predictions using CNN (Model 4).}
	\label{tab:cnn_hyperparams}
	\begin{tabular}{@{}llc@{}}
		\toprule
		\textbf{Parameter} & \textbf{Value} \\
		\midrule
		\textbf{Input configuration} & \\
		Input channels 		& 6 \\
		Slice size 			& 1000 samples \\
		Slice shift 		& 200 samples \\
		\arrayrulecolor{gray!50}
		\midrule
		\textbf{Network architecture} & \\
		Architecture 					&  Convolutional Neural Network \\
		Convolutional layers 			& 6 \\
		Convolutional layer sizes		& 64, 64, 128, 128, 256, 256 \\
		Kernel sizes 					& 5, 5, 5, 5, 5, 5 \\
		Fully connected layers 		& 1 \\
		Fully connected layer sizes 	& 4096 \\
		Activation function 			& ReLU \\
		Batch normalisation 			& True \\
		\midrule
		\textbf{Training procedure} & \\
		Optimizer 						& Adam \\
		Criterion 						& MSELoss \\
		Batch size 						& 32 \\
		Initial learning rate 			& $1 \times 10^{-3}$ \\
		Learning rate scheduler 		& step=10, gamma=0.1 \\
		Maximum epochs 					& 60 \\
		Patience 						& 15 \\
		\arrayrulecolor{black}
		\bottomrule
	\end{tabular}
\end{table}

\subsubsection{Predictive analytics}

The predictive performance of the models was evaluated using the experimental dataset described in \Cref{system_description}. The dataset shows an imbalanced distribution of impact energies, with a higher concentration of samples at lower energy levels and a lower concentration at higher levels. This distribution reflects operational realities in structural health monitoring, where low-energy impacts are more frequent. Consequently, the models are likely to be better trained and more accurate within the low-energy range that dominates real-world applications, while their reliability at higher energies may be reduced due to limited data availability. As a result, average accuracy metrics may be slightly overestimated, as model performance in the majority (low-energy) region dominates the overall evaluation. Nonetheless, this limitation is inherent in the nature of available data in real-world monitoring and affects all models similarly, thereby preserving the validity of the comparative analysis presented below.

The comparative results are summarised in \Cref{fig:input_space_validation_coupon}, which visualises both residual patterns and prediction behaviour across input space configurations. CNN (Model 4) processes each measurement using multiple overlapping sliding windows, yielding several predictions per sample. Accordingly, whiskers show the full range of predictions across these windows and indicate uncertainty, while red markers show the median predicted energy for each true energy level. Residuals for the CNN were computed from these median predictions, which represent the central estimate across all window outputs for each impact test sample. In contrast, FCNN models (Models 1--3) produce a single prediction per sample, and therefore, no whiskers are shown. The bar plot in the same figure highlights the uneven distribution of impact energies.
\begin{figure}[htbp]
	\centering
    \includegraphics{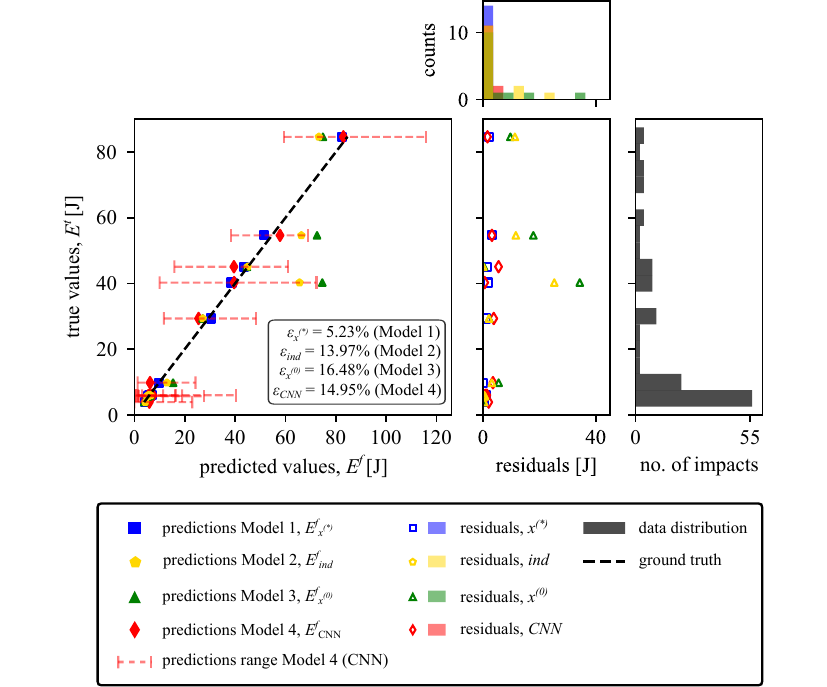}
    \caption{Comparison of impact energy predictions using physics-inspired energy indicators $x^{(*)}$, correlation-based features $x^{\text{ind}}$, and the full candidate set $x^{(0)}$ within a FCNN, and a black-box CNN.}
    \label{fig:input_space_validation_coupon}
\end{figure}

Among the assessed methodologies, the physics-inspired FCNN showed the lowest Mean Absolute Percentage Error (MAPE) at $\epsilon_{x^{(*)}} = 5.23\%$. This approach outperformed the independent features model ($\epsilon_{ind} = 13.97\%$), the CNN ($\epsilon_{CNN} = 14.95\%$), and the FCNN trained on the complete candidate features set ($\epsilon_{x^{(0)}} = 16.48\%$). 

This performance ranking is further supported by residual analysis. The physics-inspired energy indicator model shows residuals closely clustered around zero, demonstrating consistent accuracy across the entire energy range. In contrast, both alternative FCNN configurations exhibit higher residual values resulting from the incorporation of unrefined features. The latter limitation hinders the ability of the model to isolate relevant patterns, particularly as non-linearities increase and noise is introduced during data augmentation. Although correlation-based feature selection (Model 2) mitigates redundancy and yields moderate improvements, it fails to ensure robustness or sensitivity to impact energy, ultimately reducing reliability.

The CNN exhibited a distinct prediction pattern: individual windows often produced inaccurate estimates, yet aggregating their outputs through median values yielded a reliable overall prediction. In this context, prediction noise across all window outputs increased in regions with limited training data, particularly above 10~J, where the whiskers in the scatter plot of \Cref{fig:input_space_validation_coupon} broadened considerably, indicating greater variability among individual window predictions. In contrast, median outputs aligned well with the ground truth, as the median reduced the influence of outliers and provided a stable central estimate. 

The results aforementioned concern accuracy alone, which here measures how well each model estimates the severity of an impact. That measure is conclusive for Models 2 and 3, but it leaves the comparison with the CNN open, since median predictions of the CNN model achieve similar accuracy. The assessment, therefore, extends to what each input space provides beyond the estimate. Three criteria guide this discussion, all concerning the quantities a model takes as input. First, an estimate should stay traceable to interpretable quantities, since a diagnostic tool gains reliability when its output can be linked to the input that produced it~\citep{Elenchezhian2021}. Second, those quantities should remain inspectable against expected behaviour, since faults in the sensing system distort an estimate without leaving a visible sign~\citep{Wang2024}. Third, the quantities should stay general enough to serve further use in the monitoring workflow~\citep{Hunt2001, DeLuca2023}. The proposed energy indicators meet all criteria, while the CNN meets only the accuracy requirement.

The definitions of the retained descriptors establish traceability. Each indicator has an explicit formulation and identifies a characteristic of the measured response. For instance, \Cref{tab:feature_extraction} associates each indicator with the mechanisms by which it may relate to the impact event, and \Cref{fig:EDA_boxplot} confirms the corresponding trend across the energy range. The model therefore reports each estimate together with the quantities that produced it. The CNN forms its own representation while training. That representation remains internal to the network, and no predefined quantity explains its output.

An additional argument follows from the composition of the retained set. The final physics-inspired indicators cover the time, frequency and time-frequency domains. The time domain tracks the amplitude and energy released during the event; the frequency domain identifies redistribution across spectral bands; and the time-frequency domain retains the transient content of an impact. Each domain therefore carries complementary information. In contrast, the CNN model takes the raw responses, and no part of its objective rewards such a multi-domain balance.

Furthermore, the proposed indicators support a direct check of whether an estimate is produced from signals that still behave as observed during selection. This is especially useful when operation lacks ground truth and faulty sensors can corrupt SHM conclusions, producing incorrect predictions that are not recognisable from the output alone~\citep{Wang2024}. The CNN offers no comparable route, since its feature maps do not expose a documented link to the structural dynamics. Explainability methods for black-box models offer only a partial substitute, focusing mainly on characterising how a prediction was reached, rather than on input-level traceability. Meaningful indicators consequently retain a clearer advantage on checking whether an estimate remains grounded in the excitation mechanisms it describes.

Beyond this check, the indicators serve purposes outside energy estimation. Because each descriptor is tied to a specific mechanism, the same descriptors carry information relevant to severity-aware decisions, damage characterisation, and maintenance actions, the other stages of the impact identification pipeline~\citep{Broer2021}. One extraction stage, therefore, serves as a shared interface for several stages of the pipeline. The representation learned by the CNN is specific to the single task on which the network was trained, and each subsequent stage would need its own dedicated network.

The move to another structure benefits from the same generality. Definitions that hold independently of the panel and sensor layout make the workflow easier to transfer to a different component, even though a new structure would still require its own training data, or at least a fine-tuning step. The feature maps of the CNN offer no comparable starting point, since their meaning holds only inside the network that produced them.

Common definitions also connect measurements to simulation. The finite element model of the panel returns the quantities the indicators describe, so simulated and measured responses can be compared on the same terms, thereby supporting validation of the finite element model. This validation matters because simulated data are an emerging strategy for augmenting scarce experimental datasets in monitoring studies~\citep{Kudela2020, Ferreira2024}, but only if the simulation itself is sufficiently accurate. The internal features of the CNN cannot be compared with parameters in the finite element model, so this validation route remains closed to it.

The two routes also differ in where the modelling effort goes, but it cannot be argued that one outperforms the other on this point. The proposed model takes seven inputs into a compact network, whereas the CNN operates on full-response windows, so the design choices differ in both kind and number. Defining the candidate descriptors and screening them for sensitivity, redundancy and robustness takes time at the start. That time produces a documented input space, where every choice follows from expectations on structural dynamics and can be revisited later. The CNN moves that effort to depth, filter size and window length, which the user can report but cannot derive from the impact problem, and the full response windows raise the computational cost of training and of every subsequent estimate.

Overall, a compact set of energy indicators, selected through a structured workflow guided by domain knowledge, delivers accurate estimates while keeping each estimate traceable, inspectable, and available to later stages of the pipeline. The workflow also provides a route for integrating structural dynamics principles into data-driven methodologies, paving the way for impact energy assessments that an engineer can follow with high transparency.

\section{Concluding remarks}
\label{sec:conclusion}

The proposed method demonstrates that guiding feature design with domain knowledge enables more reliable estimation of impact energy, particularly under measurement constraints and limited data availability. Its main contribution is a structured selection procedure that explicitly defines the input space, reducing dependence on large training sets and delivering consistently higher predictive accuracy than models trained on unstructured data or conventional signal metrics.

Descriptors are extracted from the time, frequency, and time–frequency domains and evaluated using a composite scoring procedure that balances relevance and stability in the presence of noise. The scores quantify descriptor performance, allow transparent comparison, and guide the retention of the most informative indicators. This process yields a compact and meaningful input space that preserves interpretability while supporting accurate energy estimation.

Two aspects of this design deserve emphasis. Each retained descriptor can be related to the aspect of the response it describes, so the selection remains traceable. The scores also characterise the input space: a low importance score indicates limited contribution to the variance, and a low robustness score indicates sensitivity to noise, highlighting where the input space is weak and which descriptors would need reconsideration.

Despite notable strengths and strong predictive performance, certain limitations must be acknowledged. The magnitude of measurement noise and additional structural non-linearities can influence the stability of predictions. Furthermore, reliance on expert judgement for feature definition may introduce subjectivity when deciding which mechanisms to represent or where to set relevance thresholds, particularly in complex or poorly characterised systems.

To address these limitations, future work should extend the procedure to more complex structural configurations and enhance the integration of prior knowledge. In this way, a natural next step is to incorporate governing equations or physical constraints directly into the network architecture and learning process, extending the present approach from the input space to the model itself, thereby improving extrapolation beyond the training domain. Multi-fidelity feature extraction offers a further direction, in which features derived from simulation data complement experimental observations and improve data efficiency when testing is costly or unavailable. Together, these developments would improve generalisation and support scalability in impact energy estimation.

In summary, the proposed methodology provides a robust foundation for accurate, interpretable, and scalable models in impact estimation. Placing prior knowledge at the feature level maintains performance under practical data constraints, with clear potential for deployment in monitoring and maintenance workflows.


\ack{This work is part of the PrimaVera Project, which is partly financed by the Dutch Research Council (NWO) under grant agreement NWA.1160.18.238.}

\bibliographystyle{unsrtnat} 
\bibliography{references}

\end{document}